\documentclass{article}

\usepackage{arxiv}

\usepackage[utf8]{inputenc} 
\usepackage[T1]{fontenc}    
\usepackage{hyperref}       
\usepackage{url}            
\usepackage{booktabs}       
\usepackage{amsfonts}       
\usepackage{nicefrac}       
\usepackage{microtype}      
\usepackage{lipsum}
\usepackage{natbib}

\usepackage{amssymb}
\usepackage{latexsym}
\usepackage{float}

\usepackage{adjustbox}
\usepackage{threeparttable}

\title{Modal features for image texture classification}

\author{
  Thomas Lacombe\thanks{© 2020. This manuscript version is made available under the CC-BY-NC-ND 4.0 license \href{http://creativecommons.org/licenses/by-nc-nd/4.0/}{http://creativecommons.org/licenses/by-nc-nd/4.0/}} \\
  Univ. Savoie Mont Blanc\\
  SYMME\\
  F-74000 Annecy, France \\
  \texttt{thomas.lacombe4@gmail.com} \\
   \And
  Hugues Favreliere\\
  Univ. Savoie Mont Blanc\\
  SYMME\\
  F-74000 Annecy, France \\
  \texttt{hugues.favreliere@univ-smb.fr} \\
   \And
  Maurice Pillet\\
  Univ. Savoie Mont Blanc\\
  SYMME\\
  F-74000 Annecy, France \\
  \texttt{maurice.pillet@univ-smb.fr} \\
}

\begin{document}
\maketitle

\begin{abstract}
Feature extraction is a key step in image processing for pattern recognition and machine learning processes. Its purpose lies in reducing the dimensionality of the input data through the computing of features which accurately describe the original information. In this article, a new feature extraction method based on Discrete Modal Decomposition (DMD) is introduced, to extend the group of space and frequency based features. These new features are called modal features. Initially aiming to decompose a signal into a modal basis built from a vibration mechanics problem, the DMD projection is applied to images in order to extract modal features with two approaches. The first one, called full scale DMD, consists in exploiting directly the decomposition resulting coordinates as features. The second one, called filtering DMD, consists in using the DMD modes as filters to obtain features through a local transformation process. Experiments are performed on image texture classification tasks including several widely used data bases, compared to several classic feature extraction methods. We show that the DMD approach achieves good classification performances, comparable to the state of the art techniques, with a lower extraction time.
\end{abstract}


\section{Introduction}
Raster digital images are composed of a finite set of digital values with several dimensions (height, width, depth, time or other). In the two-dimensional space case, the most common one, these digital values are called pixels and contain a fixed number related to the color and intensity information of the scene. The amount of computing space the raster image is taking up is directly linked to its number of pixels, but also to the chosen color representation (gray-scale, RGB, HSV, etc.).

For a storage purpose, the raster image $I$ is encoded and represented with a raster file format, which aims reduce the image size without losing too much of the information it contains. Several file formats are used (JPEG, GIF, PNG, etc.) to compress the images thanks to different feature extraction techniques (i.e. Discrete Cosine Transform for JPEG).

For pattern recognition and machine learning purposes, the information contained in raster images is most of the time too complex in terms of interpretation and computing space. Before getting to the classification step, these processes are then using a feature extraction step to facilitate the learning and the generalization of the information. The reduction of information by feature extraction is therefore a major issue for the compression and classification of raster images. In this article, we decide to focus on feature extraction for classification tasks. 

More precisely, feature extraction aims to extract features from raster images, allowing to characterize and discriminate the image content with an optimal amount of information. These features can take the form of a single value or a vector of values. The feature extraction relies on a set of methods that are usually found in literature~\citep{tuceryanetal1993},~\citep{Porebski2009} as clustered into four major groups~:
\begin{itemize}
\item Geometrical approaches characterized by geometrical features,
\item Statistical or distribution approaches characterized by statistical features,
\item Filtering or transformation approaches characterized by space and frequency features,
\item Model approaches where features are based on texture spatial modelling.
\end{itemize}
One can note that the distribution of methods in these items is not strict and that some of them may belong to several groups. Each type of approach also allows to analyze a specific type of information contained in an image.

This study falls perfectly within the continuity of these description methods. Indeed, we propose the use of an original technique, based on Discrete Modal Decomposition (DMD), to extract discriminant features from raster images. The features extracted with this technique supplement and extend the group of space and frequency features.

The paper is organized as follows. In section~\ref{secattributs}, we briefly present several existing space and frequency features and their main uses in the literature. In section~\ref{secfeature}, we then introduce the genesis, the principles and the use of DMD to extract features, called modal features, from raster digital images. Finally in section~\ref{secclassif}, we propose to evaluate the relevance of these features to classify different images textures, from the VisTex database~\citep{VisTex2000}, the DTD database~\citep{cimpoi14describing}, the SIPI database~\citep{weber1997usc} and the Outex database~\citep{ojala2002outex}. We compare our technique with Haralick features from cooccurrence matrices, LBP features, Hog features and DCT features.

\section{Space and frequency features}
\label{secattributs}
Space and frequency features are mainly defined to characterize the textures contained in grayscale images. Gray levels are quantifying the luminance of the scene, which describes the amount of light that passes through, is emitted or reflected from the scene. The spacial arrangement at different scales of these gray levels characterize the texture of the image. The existence of more or less regular spatial patterns leads to a visual sensation of a contrasted, coarse, fine, smooth, granular, regular, irregular, etc. texture. Space and frequency features aim to characterize the image texture on the spatial and/or frequency domains. 

These features are usually generated by filtering or transformation approaches. They can be separated into three groups as shown by Porebski~\citep{Porebski2009}~:
\begin{itemize}
\item Spatial approach,
\item Frequency approach,
\item Space and frequency approach.
\end{itemize}
We propose in the following subsections to briefly describe the main methods of filtering or transforming in each of those three groups.

\subsection{Spatial approach}
Spatial transformation approaches can be summarized mainly in the use of filters for edge and form detection described by Tuceryan and Jain~\citep{tuceryanetal1993}. These filters are very similar to those used by the geometric features approaches. Among the broadly used methods, one can find the filters of Sobel, Prewitt, Canny or the operators of Roberts and Laplacien. The masks associated with these different filters and operators are presented in Table~\ref{tab1}.

\begin{table}[H] 
\caption{Examples of masks associated with different types of filters and operators. From left to right~: Laplacien, Prewitt, Sobel}
\centering
$\left[
\begin{array}{l l l}
-1&-1&-1 \\
-1&8&-1 \\
-1&-1&-1 
\end{array}
\right]$
$\left[
\begin{array}{c c c}
-1&0&-1 \\
-1&0&-1 \\
-1&0&-1 
\end{array}
\right]$
$\left[
\begin{array}{c c c}
-1&-2&-1 \\
0&0&0 \\
-1&-2&-1 
\end{array}
\right]$
\label{tab1}
\end{table}

\subsection{Frequency approach}
In the frequency domain, the main approaches used to describe the images are the Fourier transform and the Discrete Cosine Transform (DCT). These approaches allow to extract the image characteristics, related to the texture of these images and providing a texture description exclusively in the frequency domain. The application of these transformations allow to obtain coefficients providing frequency information about the image content.

Indeed, they are especially suited to describe images containing periodic element structures, which is for example the case of coarse texture images. 
The high frequencies are reserved for local variations of the gray level values or the color components of the pixels, thus all the information of the image can easily be represented with very few coefficients, corresponding to the low frequencies. Drimbarean uses DCT method on gray level images and its extension to color to characterize the image textures from the VisTex database~\citep{Drimbarean2001}. He concludes that the extracted features from the DCT approach allow to obtain the best rate of image texture classification, whether with a gray level or color coding, compared with the classification carried out with the extracted features from the Gabor transformation or the Haralick features extracted from the co-occurrence matrices.

\subsection{Space and frequency approach}
Finally, there are other features which describe the texture in both the space and frequency domains. Among the most used space and frequency methods are the Gabor transform and the wavelet transform. We propose to briefly present these ones in the following subsection.

\textit{Gabor transform} \\
We have previously specified that the use of the Fourier transform allows to characterize the frequency content of the images. A solution to include a spatial description of the image content is to apply the short-time Fourier transform, which principle consists in using the Fourier transform in an moving observation window.
The choice of the window size and the displacement pitch depends on the spatial patterns of the textures to be analyzed. There are various types of observation windows: the rectangular window, the triangular window, the Welch window, the Hamming window, the Hann window and the Gaussian window~\citep{deCoulon1998}. When the latter is applied, the transform is therefore called the Gabor transform.

The coefficients are calculated at each step of the sliding window. The extracted features are then often obtained by calculating the energy, the entropy or the variance of these coefficients for each window and for a given filter. Palm and Lehmann~\citep{Palm2002} show that the Gabor transform is competitive at describing image textures and these texture feature extraction from the color images significantly improve the image classification of the VisTex database compared to the use of these features defined in gray level.

\textit{Wavelet transform} \\
The Gabor transformation is applied with a fixed window size, which can be one of its limits because some textures can be characterized according to various scales. In order to overcome this limit, the wavelet transform is based on a multiscale analysis of the images, and uses observation windows with various sizes.

Arivazghan uses features extracted trough the wavelet transform to classify the texture images of the VisTex database~\citep{Arivazghan2005}. In the same way, Sengur uses this kind of features, especially through the calculation of the energy of the various filters, to highlight their contribution to the texture classification of the same database, focusing especially on the color textures~\citep{Sengur2008}. This transformation has also become famous with its application for image compression with JPEG2000 format~\citep{Antonini1992}.

\section{Feature extraction by Discrete Modal Decomposition}
\label{secfeature}
In this section, we detail the principles of a new spatial and frequency feature extraction technique based on Discrete Modal Decomposition.
With the initial goal of automatically generating a 3D model for Computer Aided Design, Pentland~\citep{Pentland1990} chose to use dynamic behavior of the object, i.e. its natural modes of vibration to describe its shape. Based upon this idea, the Discrete Modal Decomposition consists in decomposing a signal within a spectral basis built from eigen modes. 

Similarly to the Discrete Fourier Transform or the Discrete Cosine Transform, this projection method allows to make the projection of the signal into an eigen basis built from structural dynamics. This eigen basis is defined by its eigen vectors, called modal vectors, which explains the name chosen for the projection method.

The DMD is classified as a descriptive approach, using a description basis which is not related to the solution of a given problem. This decomposition method has allowed to parameterize geometric shape deviations in the geometrical tolerancing domain~\citep{Samper2007},~\citep{Favreliere2009}.
It was then extended by Le Go{\"\i}c to characterize and filter geometric deviations of higher frequencies, including undulations and roughness~\citep{LeGoic2011} and compare its performance with standard methods, to correlate roughness measurements with their tribological behaviors~\citep{LeGoic2016}. More recently, in the field of surface appearance modeling, DMD has significantly improved the reflectance function approximation for a more realistic rendering and highlighting appearance anomalies~\citep{Pitard2017}.

\subsection{Modal basis}
\begin{figure}[!t]
\centering
\includegraphics[scale=.27]{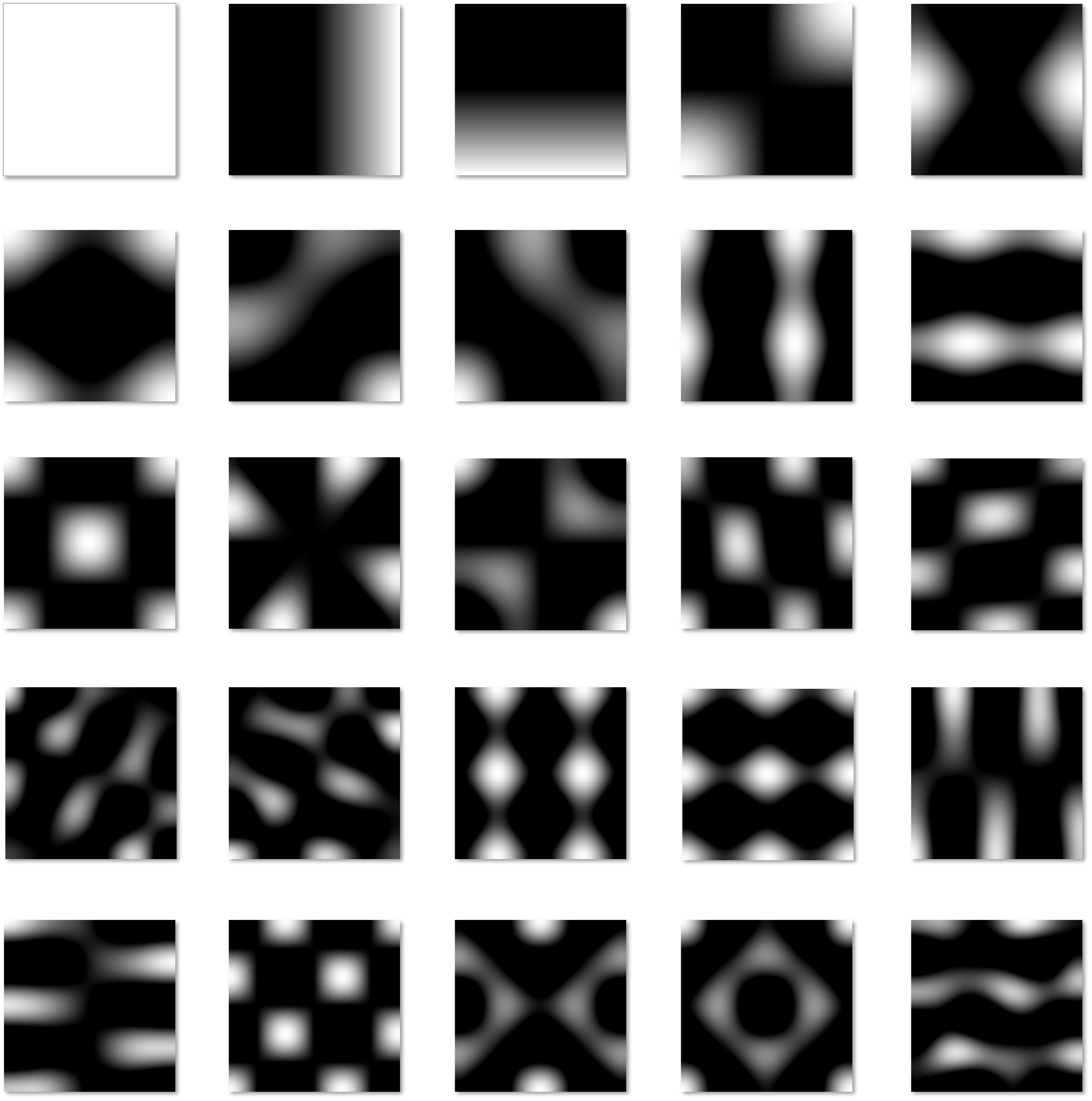}
\caption{Modal projection basis. Only the first 25 modes are presented.}
\label{figmodalbasis}
\end{figure}

A rectangular domain designates the solid structure under investigation. The associated geometry (a square plate in the proposed study) leads to the definition of the following dynamic structural problem (Eq.~\ref{eqdyna}):
\begin{equation}
M\cdot\ddot{q}+K\cdot q=0 \textrm{   with   } q=q(x,y,t)
\label{eqdyna}
\end{equation}
where $M$ and $K$ stand for the mass and the stiffness matrices respectively. Under such formalism, $q(x,y,t)$ stands for the displacements which characterizes the modal shapes. Such a problem classically gives a frequency based solution (Eq.~\ref{eqsolu}):
\begin{equation}
q(x,y,t)=\sum\limits_{i=1}^{+\infty}Q_i(x,y).\cos(\omega_{i}t)
\label{eqsolu}
\end{equation}
where $Q_i$ is the magnitude vector associated with the pulsation $\omega_i$. Hence, in the finite dimension framework, the eigen modes defined by $\left(Q_i,i\right)$ are determined by solving the following linear system (Eq.~\ref{eqline}):
\begin{equation}
\left(M^{-1}K-\frac{1}{\omega_{i}}\mathbb{I}\right)Q_i=0
\label{eqline}
\end{equation}
where $\mathbb{I}$ is the identity matrix and $M^{-1}K$ is assured to be diagonalizable. The discrete solution is computed by using Finite Element Analysis (FEA) and provides the dynamic modal basis $Q=\left(Q_1, Q_2\dots\right)$. As an illustration, the first 25 modal vectors are plotted in Fig.~\ref{figmodalbasis}. 

\subsection{Decomposition operation}
The decomposition operation is carried by projecting the image pixel values onto the eigen modes, built from dynamics and namely the modal basis. However, the non orthonormality of $Q$ does not allow the use of the classical projector $P_Q = QQ^T$. Indeed, the use of the dual basis $Q^* = \left(Q^{T}Q\right)^{-1}Q^T$ is required. An infinite norm is given to the modal vectors such as $\|Q_i\|_{\infty}=1$. Thus, the set of modal coordinates $\lambda_i$, modal features that we use in the proposed study, resulting from the projection of the image pixel values (denoted $P_V$) within a non-orthonormal basis is given as follows (Eq.~\ref{eqproj}):
\begin{equation}
\lambda=\left(Q^TQ\right)^{-1}Q^T.P_V
\label{eqproj}
\end{equation}
Thus, the image pixel values can be expressed as the sum of linear combination of the modal vectors and the decomposition residual (Eq~\ref{eqcomb}):
\begin{equation}
P_V=\sum\limits_{i=1}^{N_q}\lambda_{i}Q_i+\epsilon\left(N_q\right)
\label{eqcomb}
\end{equation}

with $\left \{
\begin{array}{l c l}
Q_{i} & : & modal \ vectors \ composing \ the \ modal \ basis	\\
\lambda_{i} & : & modal \ coordinates \ = \ modal \ features	\\ 
N_{q} & : & number \ of \ modes \ of \ decomposition
\end{array}
\right.$

\begin{figure}[H]
\centering
\includegraphics[scale=.3]{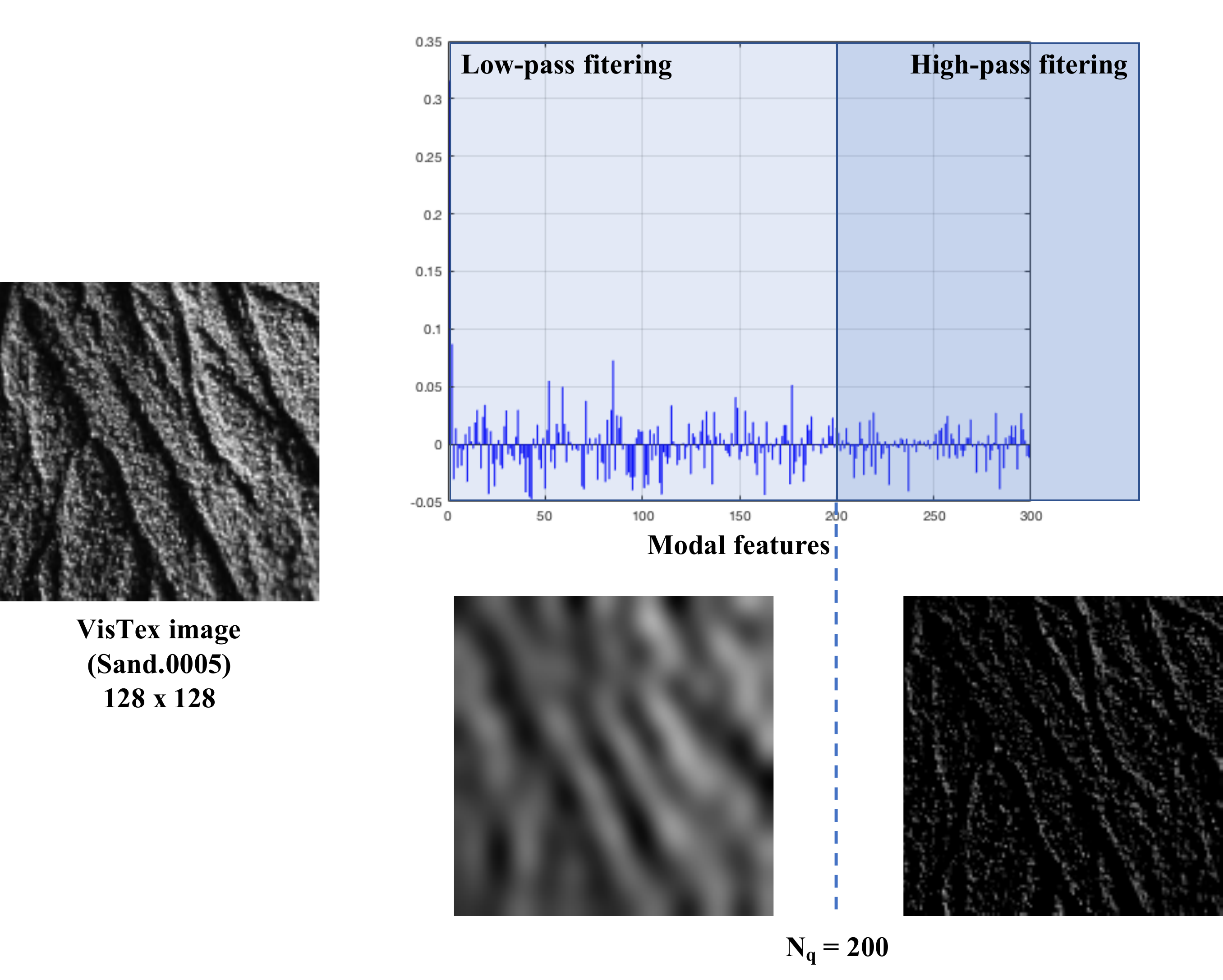}
\caption{DMD filtering applied on an image of VisTex database namely (Sand.0005)}
\label{figmodalfilter}
\end{figure}

\subsection{Rotation invariant modes}
Most of the time, image texture patterns are likely to vary in terms of orientation without changing the texture itself. These orientation changes must then not be considered as a variation of the texture itself when it comes to a classification task. To avoid this, it is possible to use rotation invariant properties of the modal basis to perform the classification task.

The natural modal projection basis has two types of modes~\citep{Favreliere2009}, identifiable by their geometrical properties and defined by~\citep{pitard2016metrologie}~:
\begin{itemize}
    \item simple modes~: modes with orthogonal symmetry, therefore their shapes are invariant by rotation,
    \item congruent modes~: couple of modes naturally generated with an angular shift.
\end{itemize}
Pitard proposes to use these properties to enhance the angular reflectance function and to demonstrate its relevance to describe the appearance of the inspected surfaces in the visual quality inspection field~\citep{pitard2017reflectance}. Following this aim, he suggests to transform the two modal coordinates $\lambda_i$ and $\lambda_{i+1}$ of the couple of congruent modes $Q_i$ and $Q_{i+1}$ resulting from the projection operation (Equation~\ref{eqproj}) into a single amplitude value $\lambda_j^{'}$ and a single phase value $\phi_j^{'}$. The resulting amplitude value is then rotation invariant and expressed~:
\begin{equation}
\lambda_j^{'}=\sqrt{\lambda_i^2+\lambda_{i+1}^2}
\label{eqinv}
\end{equation}
This new setting ($\lambda_j^{'}$ and $\phi_j^{'}$) of the congruent modes allowed him to estimate robust salience maps linked to the local visual appearance behaviour of surfaces on the scene.

\subsection{Multiscale analysis application}
Initially implemented to characterize form variations~\citep{Samper2007},~\citep{Favreliere2009} on the geometrical tolerancing, this method has been generalized to waviness and roughness by varying the width of the analysis window~\citep{LeGoic2011},~\citep{Grandjean2012}.

\begin{figure*}[!t]
\centering
\includegraphics[scale=.50]{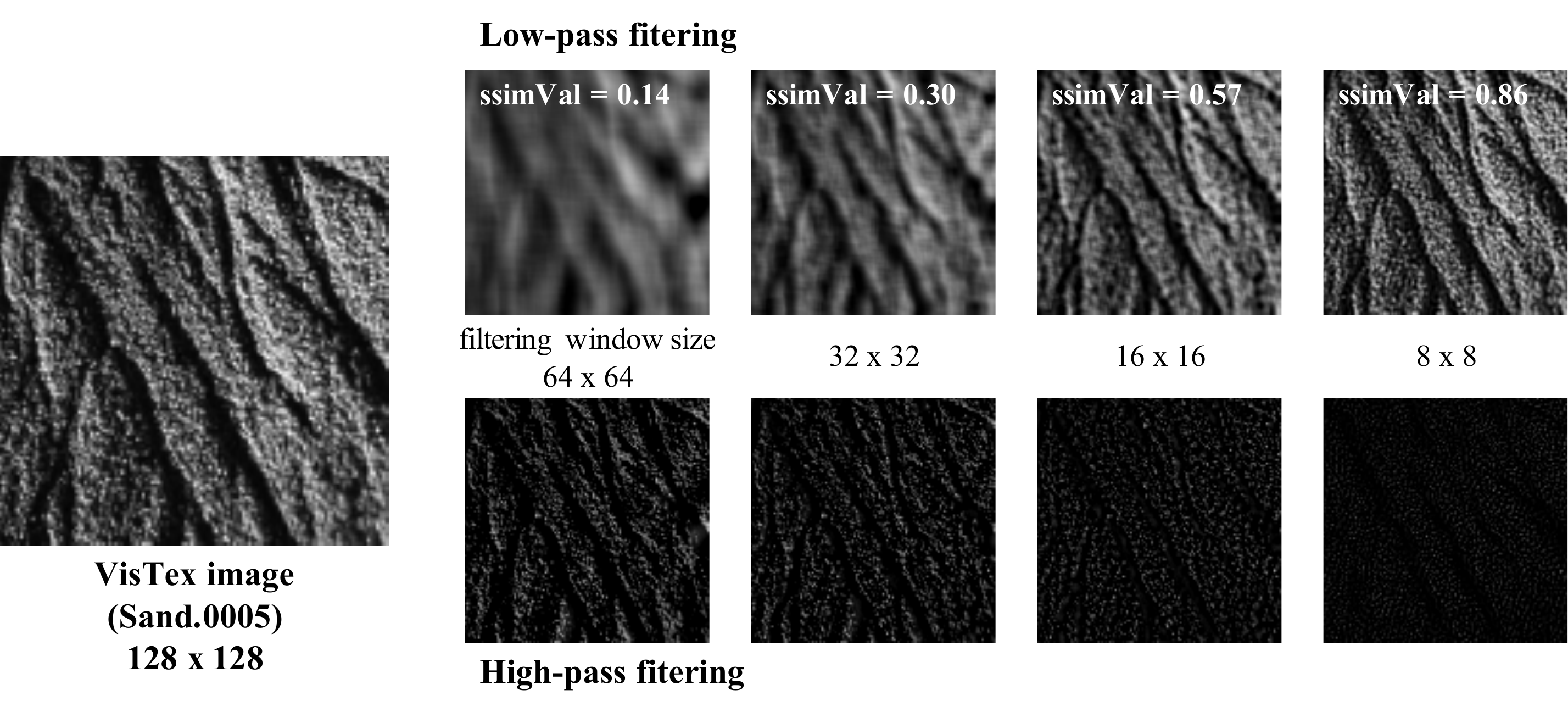}
\caption{DMD multiscale analysis applied on an image of VisTex database namely (Sand.0005)}
\label{figmodalfilterg}
\end{figure*}

Once the projection operation is computed, the result can be viewed as an modal features amplitude spectrum. It enables separate the low frequency and the high frequency (and also a frequency band) content of the image texture, rebuilding the image from a part of the modal features spectrum. This general principle is illustrated in Figure~\ref{figmodalfilter}. As an illustration, Figure~\ref{figmodalfilterg} shows 4 low-pass and high-pass content resulting from the application of the DMD multiscale analysis technique on the same VisTex image (Sand.0005) with $N_q=20$. 

\section{Image textures classification}
\label{secclassif}

To evaluate the relevance of the DMD applied as a feature extraction method, its performance on an image textures classification task has been studied, in parallel of a few others techniques broadly used for texture and image description : Haralick's cooccurrence features, Local Binary Patterns (LBP), \textbf{Histograms of Oriented Gradients (Hog)} and DCT features.
In this section, we briefly explain the principles of the different features extraction methods used in the study before outlining the experimental set-up and reporting the results. 

\subsection{Feature extraction}

This section briefly describe the feature extraction methods we used to conduct this classification study. 

\subsubsection{Cooccurrence matrix and Haralick features}

The cooccurrence matrix, classified as a statistical feature extraction method, is used to characterize the texture of an image based on its pixel values distribution. The matrix is computed by counting all pairs of pixels with gray levels $i$ and $j$ separated by a distance $d$ in a given direction $\theta$. The matrix is often calculated by accumulating the values of cooccurrences in several directions to obtain a rotation invariant description. In this study, the matrices are calculated with $\theta = \{0,\pi,\frac{\pi}{2}, \frac{3\pi}{4}\}$ and $d=1$.
Once calculated, the cooccurrence matrices are usually reduced by extracting a set of features. The most popular ones have been proposed by Haralick~\citep{haralick1973textural} and have been widely used in texture analysis. 

We give the equations of the first 5 features as an illustration :
\begin{enumerate}
\item Angular Second Moment
\begin{equation}
    h_1 = \sum_i \sum_j {p(i,j)}^2
\end{equation}
with $p(i,j)$ the probability that a pixel with value $i$ will be found adjacent to a pixel of value $j$.
\item Contrast
\begin{equation}
    h_2 = \sum_{n=0} n^2 \left\{\sum_{i=1}^{N_g} \sum_{j=1}^{N_g} {p(i,j)}\right\}, |i-j|=n
\end{equation}
\item Correlation
\begin{equation}
    h_3 = \frac{\sum_i \sum_j (ij)p(i,j)-\mu_x \mu_y}{\sigma_x \sigma_y}
\end{equation}
 with $\mu_x$, $\mu_y$, $\sigma_x$ and $\sigma_y$ the means and standard deviations of $p_x$ and $p_y$, the marginal-probability matrices obtained by summing the rows or columns of $p(i,j)$.
\item Sum of Quares : Variance
\begin{equation}
    h_4 = \sum_i \sum_j {i-\mu}^2 p(i,j)
\end{equation}
\item Inverse Difference Moment
\begin{equation}
    h_5 = \sum_i \sum_j \frac{1}{1-(i-j)^2} p(i,j)
\end{equation}
\end{enumerate}
The total of the 14 features equations can be found in Haralick's original publication~\citep{haralick1973textural}.

\subsubsection{Local Binary Patterns (LBP)}

LBP, also classified as a statistical feature extraction method, are used to characterize the texture of an image based on a comparison of the luminance level of each pixel with its neighbors~\citep{ojala1996comparative}. Each pixel value is replaced by a weighted sum described in Equation \ref{lbp_sum}.

\begin{equation}
    LBP_{P,R}(x_c,y_c) = \sum_{p=0}^{P-1} 2^p \delta (g_p - g_c)
    \label{lbp_sum}
\end{equation}
where $P$ is the number of pixels to consider around pixel $(x_c,y_c)$ , R is the radius defining the neighborhood, $g_c$ is the gray level of pixel $(x_c,y_c)$ , and $g_p$ is the grey level of the $p$ neighbor. 

The feature is constructed as the histogram of the LBP values over the image.

\subsubsection{Histograms of Oriented Gradients (Hog)}

Hog descriptors can be used to characterize the distribution of intensity gradients and therefore edge directions in an image. The image is spitted in cells for which a gradient histogram is computed. The gradient is obtained by applying a mask to filter the cell. The final descriptor is build by concatenating all the histograms generated for each cell.

This method is commonly used to detect objects in images, such as humans detection~\citep{dalal2005histograms}. It is not designed to perform especially well in texture description, but we still include it in this study to test if our approach performs better than a non specialist one for this specific task.

\subsubsection{Discrete Cosine Transform (DCT)}
DCT is part of the local linear transforms family~\citep{unser1986local}. These transforms are used to characterize image texture~\citep{ahmed1974discrete} in the frequency domain through a filter bank of relatively small size (a few pixels on each side). Each filter is designed to capture a particular signature of the local texture patterns.\\
DCT has been widely used in image feature extraction, particularly for image coding purpose with the JPEG encoding for example.\\
The DCT filter bank is built from a set of $N\times1$ basis column vectors $h_m$, computed as followed :
\begin{equation}
    h_m =
    \left\{
\begin{array}{lr}
  \frac{1}{\sqrt{N}} &if\ m = 0,\\
\sqrt{\frac{2}{N}}\cos(\frac{(2k - 1)m\pi}{2N}) &if\ m > 0.
\end{array}
\right.
\label{eq:DCTvectors}
\end{equation}

To compute the $N\times2$ filters, the vectors defined in Eq.\ref{eq:DCTvectors} are combined thanks to the outer product : 
\begin{equation}
    h_{mn} = h_mh_{n}^T
\label{eq:DCTfilters}
\end{equation}

To extract features from an image $I$, we choose to follow the same procedure than~\citep{Drimbarean2001} who defines the texture features as the variance of the filtered $M \times M$ image $I_{mn}$ by the $h_{mn}$ filter defined by Eq.\ref{eq:DCTfilters}. The filtered image $I_{mn}$ and the features $f_{mn}$ are then calculated using the following equations :
\begin{equation}
    I_{mn} = I * h_{mn}
\label{eq:convDCT}
\end{equation}
where $*$ the 2D convolution operator. 

\begin{equation}
    f_{mn} = \frac{1}{M} \sum_{x,y = 0}^{M}(I_{mn}(x,y) - \mu_{mn})^2
\label{eq:DCTfeatures}
\end{equation}
where $\mu_{mn}$ is the mean of the filtered image $I_{mn}$.

In our study, the DCT features are computed for a filter size $N=3$. In the implementation we used, also following~\citep{Drimbarean2001}, the 1D DCT vectors defined in Eq.\ref{eq:DCTvectors} are $h_0 = \{1,1,1\}$, $h_1 = \{1,0,-1\}$ and $h_2 = \{1,-2,1\}$. Using Eq.\ref{eq:DCTfilters}, a set of of nine 2D DCT filters are calculated and used to compute nine texture features according to Eq.\ref{eq:convDCT} and \ref{eq:DCTfeatures}.

\subsubsection{DMD approaches}
\label{subsubsec:DMDApproaches}

\begin{figure}[!t]
\centering
\includegraphics[scale=.28]{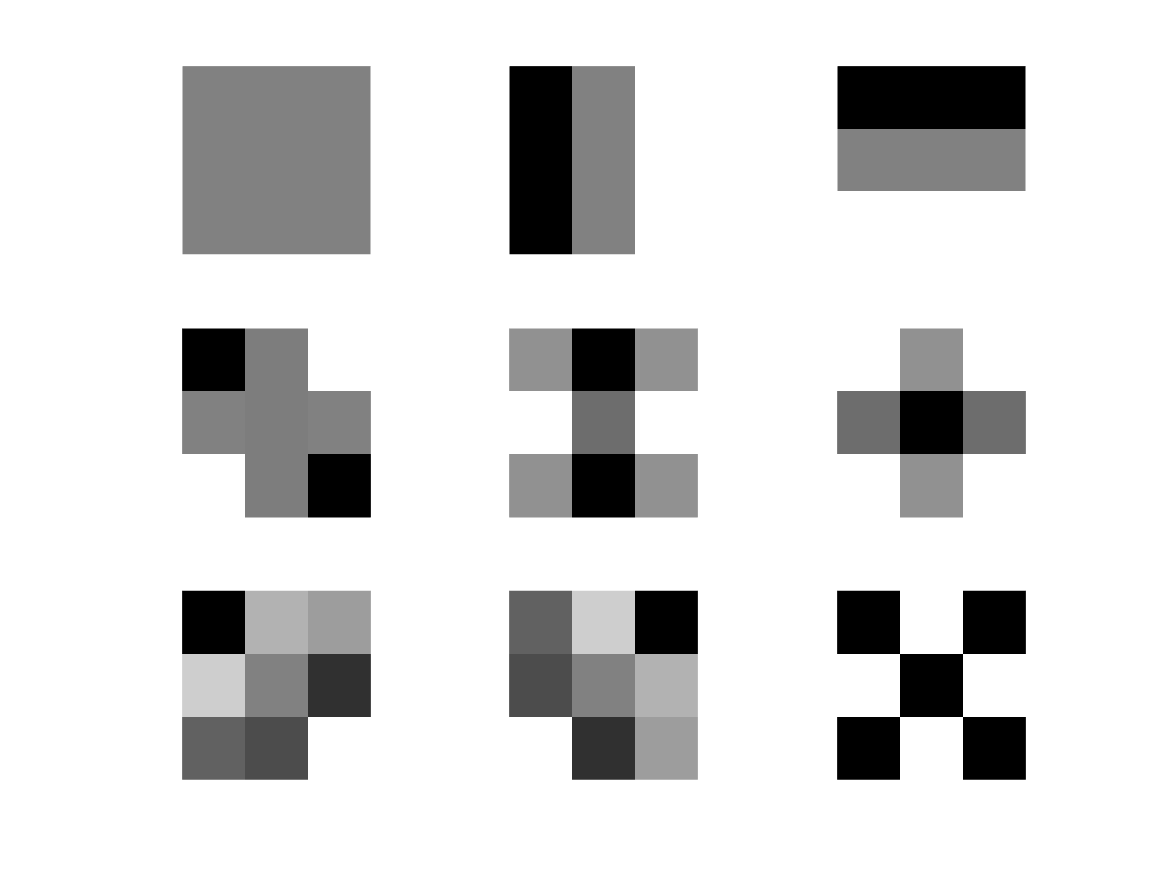}
\caption{DMD filters}
\label{imageDMDfilter}
\end{figure}

We used two approaches to extract modal features via the DMD method explained previously :  
\begin{enumerate}
    \item The Full scale DMD~: \\
    The invariant modal coordinates $\lambda_j^{'}$ obtained from the full image decomposition (Equation \ref{eqinv}) are directly used as features. 
    \item Filtering DMD~: \\
    The modal features are obtained using the same Eq.\ref{eq:convDCT} and \ref{eq:DCTfeatures} than for the DCT features, except that the filters $h_{mn}$ are different. For a filter size $N=3$, the DMD filters are the first nine modes of the projection basis. These modes are illustrated in the Fig.\ref{imageDMDfilter}.
\end{enumerate}

\subsection{Classification method}

Once the information contained in the images is reduced into a vector of features thanks to the methods presented in the previous section, the next step rely in using a classifier to evaluate the discriminatory power of the extracted features. The classification is composed of two phases : the learning and the prediction. The learning consists in feeding the algorithm with training data so it can build a model to explain the data from the extracted features. In our case, we use a supervised approach, which means that the training data is labelled by a human expert. Following this learning phase, the training algorithm is able to predict the labels of new elements, which are not part of the training data. The classification performance, or discriminatory power of the features, is usually characterized by the number of good predictions obtained from this second phase. 

There are numerous types of classification methods, and we have adopted the Support Vector Machine (SVM)~\citep{vapnik1995nature} to perform the experiments presented in this study. The detailed explanation of the SVM functioning goes beyond the scope of this article, but the elementary principle is to separate the data with hyperplans in the multidimensional space. If the original method works only for binary classification problems and linearly separable data, several developments have allowed to apply it on non linear data~\citep{boser1992training} and multi-class problems~\citep{weston1999support}. The results shown in this paper are obtained with a One-Versus-Rest SVM classifier, which enables multi-class classification.  

\subsection{Datasets}

To assess the performance of the features, 4 datasets has been used :

\subsubsection{\textit{VisTex database}}
\textit{16 textures} images picked from the 19 VisTex textures~~\citep{VisTex2000} are used, namely : Bark.0001, Brick.0000, Clouds.0000, Fabric.0001, Leaves.0010, Flowers.0000, Food.0000, Grass.0001, Metal.0000, Misc.0002, Sand.0004, Stone.0005, Tile.0007, Water.0000, Wood.0002 and WheresWaldo.0001.

\subsubsection{\textit{DTD database}}
\textit{22 textures} images from the DTD textures~~\citep{cimpoi14describing} are used, namely : banded\_0137, braided\_0078, bubbly\_0054, bumpy\_0079, cracked\_0129, crystalline\_0124, dotted\_0164, fibrous\_0193, flecked\_0126, freckled\_0159, frilly\_0121, grid\_0102, grooved\_0110, honeycombed\_0133, marbled\_0186, pleated\_0111, porous\_0174, scaly\_0222, smeared\_0132, stratified\_0162, striped\_0058,  waffled\_0136.

\subsubsection{\textit{SIPI database}}
\textit{16 textures} images from the SIPI textures~~\citep{weber1997usc} are used, namely : 1.1.01, 1.1.02, 1.1.03, 1.1.04, 1.1.05, 1.1.06, 1.1.07, 1.1.08, 1.1.09, 1.1.10, 1.1.12, 1.1.13, 1.5.02, 1.5.03, 1.5.04, 1.5.06.
We also use \textit{12 textures} from the SIPI rotated textures, namely : bark, brick, bubbles, grass, leather, pigskin, raffia, sand, straw, water, weave, wood, wool. We use orientations at 0, 30, 60, 90, 120, 150 and 200 degrees. 

For the 3 previous data bases, following the same procedure as~\citep{Drimbarean2001} to enhance the number of training and testing images, each image is randomly divided into 540 overlapped $32 \times 32$ sub-images. For each class, 30 images are used as a training set for the classifier and the 510 left are used as a testing set to evaluate the performance. 

\subsubsection{\textit{Outex database}}
Textures from 2 sets for classification prepared with the Outex database~~\citep{ojala2002outex} are used. The first, Outex\_TC\_00000, contains 480 images with 24 classes. The second, Outex\_TC\_00013 contains 1360 images with 68 classes. We don't resize the images as previously and use directly the $128 \times 128$ original images. 

\subsection{Experiments and results}

In the last part of this section, two experiments are presented to evaluate the relevance of the features extracted through the DMD. The first one focuses on the full scale DMD features and the second one on the filtering DMD features. In either case, the results are compared to a reference obtained with other methods.  

\subsubsection{Full scale DMD}

This first experiment evaluates the performance of the modal features calculated on the full scale image. 

\begin{figure}
\centering
\includegraphics[scale=.4]{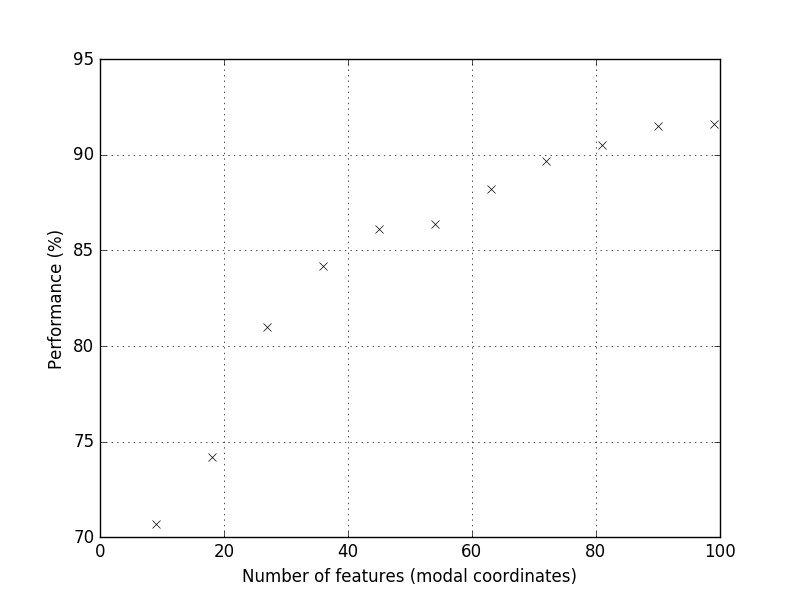}
\caption{Evolution of full scale DMD features performance as a function of the number of modal coordinates, using VisTex dataset}
\label{perfVSmodes}
\end{figure}

Before evaluating these features on the different datasets, we run an experiment to choose how many coordinates to use from the modal decomposition. The Figure \ref{perfVSmodes} shows the evolution of the performance as a function of the number of modes, up to 100 modes, based on the VisTex dataset. 
It seems that the more features are added, the more the performance increases, until a stable 91,5\% performance level from 90 features and more. The number of features to reach the performance plateau might vary from one dataset to the other, but we decide to use 100 features in the following experiment and make the assumption that they are enough to reach good performances across the different datasets.

Tables \ref{tab:resultsAcc} and \ref{tab:resultsTime} contains the SVM accuracy performances and extraction times for each method across the different datasets. The left parts correspond to the results for the full scale DMD. We notice that DMD features lead to good classification results across the different datasets, comparable to Haralick and LBP features, which are known to be especially adapted for texture classification tasks. Moreover, they always outperform Hog features, which are not specifically designed for texture description. These results demonstrate that the modal features extracted through the full scale DMD approach are relevant to classify image textures.

When it comes to extraction times results, the full scale DMD outperforms all of the 3 other techniques for all data bases. This can be explained by the fact that the full scale DMD extraction process consists in a simple dot product between the image and the dual basis. This implies a much lower computational burden than calculating the cooccurrence matrix for Haralick or the LBP image.

\begin{table}
\caption{SVM accuracy (\%) of the DMD Full Scale (FS) features against Haralick, LBP and Hog features, and Filtering DMD against DCT on several datasets}
\label{tab:resultsAcc}
\centering
\begin{adjustbox}{max width=\columnwidth}
\begin{threeparttable}
\begin{tabular}{l||c|c|c|c||c|c}
\hline
Dataset & FS DMD & Haralick & LBP & Hog & Filtering DMD & DCT \\
\hline
\hline
VisTex & 87.5 & \textbf{89.0} & 88.3 & 60.5 & \textbf{95,4} & 94,3 \\
\hline
DTD & 90.5 & 91.5 & \textbf{96.3} & 71.5 & 95.2 & \textbf{95.6} \\
\hline
SIPI & 95.0 & 97.8 & \textbf{99.8} & 92.8 & 99.1 & \textbf{99.3} \\
\hline
Outex 00 & 84.6 & 65.4 & \textbf{94.2} & 68.3 & 75.4 & \textbf{76.3} \\
\hline
Outex 13 & 72.5 & \textbf{81.3} & 68.5 & 41.5 & 75.4 & \textbf{76.2} \\
\hline
SIPI Rotated & 77.7 & \textbf{87.5} & 37.5 & 21.3 & / & / \\
\hline
\hline
Mean\tnote{a} & 86.0 & 85.0 & \textbf{89.4} & 66.9 & 88.1 & \textbf{88.3} \\
\hline
\end{tabular}\begin{tablenotes}
\small
\item[a] Mean values are calculated without taking SIPI Rotated in consideration.
\end{tablenotes}
\end{threeparttable}
\end{adjustbox}
\end{table}

\begin{table}
\caption{Extraction time (s) per image for the DMD full scale features against Haralick, LBP and Hog features , and Filtering DMD against DCT on several datasets}
\label{tab:resultsTime}
\centering
\resizebox{\columnwidth}{!}{%
\begin{tabular}{l||c|c|c|c|c||c|c}
\hline
Dataset & Image Size & FS DMD & Haralick & LBP & Hog & Filtering DMD & DCT \\
\hline
\hline
VisTex & $32\times32$ & \textbf{1.22E-05} & 1,55E-02 & 6.42E-04 & 2.90E-04 & \textbf{1.62E-04} & 1.67E-04 \\
\hline
DTD & $ 64\times64$ & \textbf{3,37E-05} & 2,26E-02 & 1,97E-03 & 1,47E-03 & \textbf{3.73E-04} &  3.88E-04 \\
\hline
SIPI & $ 128\times128$ & \textbf{8.37E-04} & 2.27E-02 & 5.39E-03  & 8.95E-03 & \textbf{1.17E-03} & \textbf{1.17E-03} \\
\hline
Outex 00 & $ 128\times128$ & \textbf{4,86E-04} & 1,36E-02 & 5,25E-03 & 7,07E-03 & \textbf{1.17E-03} & 1.23E-03 \\
\hline
Outex 13 & $ 128\times128$ & \textbf{8,96E-04}  & 9,11E-03 & 5,41E-03 & 7,57E-03 &  \textbf{1.16E-03} &  1.18E-03 \\
\hline
SIPI Rotated & $ 128\times128$ & \textbf{5.36E-04}  & 1.76E-02 & 5.28E-03 &   6,80E-03 & /  &  / \\
\hline
\hline
Mean & / & \textbf{4,67E-04} & 1,68E-02 & 3,99E-03 & 5,36E-03 & \textbf{5,62E-03} & 5.81E-03 \\
\hline
\end{tabular}%
}
\end{table}

			
\subsubsection{Filtering DMD}

This second experiment evaluates the performance of the modal features computed through the filtering DMD approach. As explained in \ref{subsubsec:DMDApproaches}, in our experiment the filtering DMD process leads to nine modal features. For comparison purpose, we also used nine DCT features, coming from Eq.\ref{eq:convDCT} and \ref{eq:DCTfeatures}.

The results illustrated in Table \ref{tab:resultsAcc} (right part) reveal that the filtering DMD features present a very good performance on the classification task, at a similar level than the DCT features, demonstrating the relevance of the filtering DMD approach to characterize image textures. It is not surprising to obtain close results with the DCT as the feature calculation method is similar for the two methods, except the base patterns for the $3\times3$ filters are different. The results are very especially good for the 3 first datasets because they contain uniform and distinct texture. Indeed, a small size of filter is particularly adapted to describe uniform textures. The textures in Outex00 and Outex13 are also uniform but there are more classes with less distinct differences between textures. This could explain the inferior classification performances on these 2 data bases.

Extraction times results in Table \ref{tab:resultsTime} (right part) show that the two methods yields very similar performances, with a slit advantage for the filtering DMD. The similar results are logical because the two methods use a similar process with a $3\times3$ moving filtering over the image.

The scope of this paper is only focused on a first investigation of the relevance of the modal features, so we only used $3\times3$ filters. This second experiment could be continued by testing different sizes of filters to see how the performance evolves. Exploring different size of filters could be particularly interesting for images with composed textures, for which a multiscale analysis would be more relevant.

\subsubsection{Rotation invariance}

We evaluate the rotation invariance performance of the full scale DMD features with the SIPI rotated dataset. The classification task includes rotated pictures of the same class at 0, 30, 60, 90, 120, 150 and 200 degrees. We compare the performance to the Haralick features, known to be rotation invariant, to the a non rotation invariant implementation of the LBP and to the Hog features, not considered as rotation invariant. We can see that if the performance of the full scale DMD features is lower to the Haralick one, it drops much less than the non invariant LBP and Hog performance. Future work will focus on investigating the rotation invariance on more datasets.

\section{Conclusions}

The aim of this study is to introduce a new feature extraction method based on the Discrete Modal Decomposition, to extend the group of the space and frequency approaches. Two classification experiments has been led to evaluate the relevance of this new approach to describe image texture, through two features extraction methods : the full scale DMD and the filtering DMD. The results illustrate the good classification performances of the two methods, competing with two other widely used approaches. In terms of extraction time, the DMD features outperforms the other approaches, especially for the full scale method. This can be interesting for application needing short computation times. 

This work opens up a lot of perspectives regarding the feature extractions methods we introduce. About the full scale DMD, it would be interesting to push further our tests by processing a feature selection step to choose what modal coordinates are the more relevant in the calculated spectrum, for a given task. Further investigation about the rotation invariance property also needs to be carried out. Future work will also focus on testing the illumination change robustness of our approach, but we cannot claim anything about it at this point. About the filtering DMD, a multiscale analysis could be performed to test the relevance of this approach on composed textures. Finally, the modal features could also be tried out for image coding tasks, to compare its performance with other commonly used formats. 

\bibliographystyle{authordate1} 
\bibliography{references}  

\begin{thebibliography}{}

\bibitem[\protect\citename{Ahmed {\em et~al.\ }\relax,
  }1974]{ahmed1974discrete}
Ahmed, N., Natarajan, T., \& Rao, K.~R. 1974.
\newblock Discrete cosine transform.
\newblock {\em IEEE transactions on Computers}, {\bf 100}(1), 90--93.

\bibitem[\protect\citename{Antonini {\em et~al.\ }\relax, }2008]{Antonini1992}
Antonini, M., Barlaud, M., Mathieu, P., \& Daubechies, I. 2008.
\newblock Image coding using wavelet transform.
\newblock {\em IEEE Transactions on image processing}, {\bf 1 (2)}, 205–220.

\bibitem[\protect\citename{Arivazhagan {\em et~al.\ }\relax,
  }2005]{Arivazghan2005}
Arivazhagan, S., Ganesan, L., \& Angayarkanni, V. 2005.
\newblock Color texture classification using wavelet transform.
\newblock {\em In Proceedings of the Sixth International Conference on
  Computational Intelligence and Multimedia Applications}, {\bf (ICCIMA)},
  315–320.

\bibitem[\protect\citename{Boser {\em et~al.\ }\relax,
  }1992]{boser1992training}
Boser, B.~E., Guyon, I.~M., \& Vapnik, V.~N. 1992.
\newblock A training algorithm for optimal margin classifiers.
\newblock {\em Pages  144--152 of:} {\em Proceedings of the fifth annual
  workshop on Computational learning theory}.
\newblock ACM.

\bibitem[\protect\citename{Cimpoi {\em et~al.\ }\relax,
  }2014]{cimpoi14describing}
Cimpoi, M., Maji, S., Kokkinos, I., Mohamed, S., , \& Vedaldi, A. 2014.
\newblock Describing Textures in the Wild.
\newblock {\em In:} {\em Proceedings of the {IEEE} Conf. on Computer Vision and
  Pattern Recognition ({CVPR})}.

\bibitem[\protect\citename{Dalal \& Triggs, }2005]{dalal2005histograms}
Dalal, Navneet, \& Triggs, Bill. 2005.
\newblock Histograms of oriented gradients for human detection.
\newblock {\em Pages  886--893 of:} {\em 2005 IEEE computer society conference
  on computer vision and pattern recognition (CVPR'05)},  vol. 1.
\newblock IEEE.

\bibitem[\protect\citename{De~Coulon, }1998]{deCoulon1998}
De~Coulon, F. 1998.
\newblock {\em Trait{\'e} et traitement des signaux (volume IV) : Th{\'e}orie
  et traitement des signaux}. Presses polytechniques et universitaires romandes
  edn.

\bibitem[\protect\citename{Drimbarean \& Whelan, }2001]{Drimbarean2001}
Drimbarean, A., \& Whelan, P.~F. 2001.
\newblock Experiments in colour texture analysis.
\newblock {\em Pattern Recognition Letters}, {\bf 22 (10)}, 1161--1167.

\bibitem[\protect\citename{Favreli{\`e}re, }2009]{Favreliere2009}
Favreli{\`e}re, H. 2009.
\newblock {\em Tol{\'e}rancement modal : de la m{\'e}trologie vers les
  sp{\'e}cifications}.
\newblock Ph.D. thesis, Universit{\'e} Savoie Mont Blanc.

\bibitem[\protect\citename{Grandjean {\em et~al.\ }\relax,
  }2012]{Grandjean2012}
Grandjean, J., Le~Go{\"\i}c, G., Favreli{\`e}re, H., Ledoux, Y., Samper, S.,
  Formosa, F., Devun, L., \& Gradel, T. 2012.
\newblock Multi-scalar analysis of hip implant components using modal
  decomposition.
\newblock {\em Measurement Science and Technology}, {\bf 23}.

\bibitem[\protect\citename{Haralick {\em et~al.\ }\relax,
  }1973]{haralick1973textural}
Haralick, R.~M., Shanmugam, K., {\em et~al.\ }\relax. 1973.
\newblock Textural features for image classification.
\newblock {\em IEEE Transactions on systems, man, and cybernetics},  610--621.

\bibitem[\protect\citename{Le~Go{\"\i}c {\em et~al.\ }\relax,
  }2011]{LeGoic2011}
Le~Go{\"\i}c, G., Favreli{\`e}re, H., Samper, S., \& Formosa, F. 2011.
\newblock Multiscale modal decomposition of primary form, waviness and
  roughness of surfaces.
\newblock {\em Scanning}, {\bf 33 (5)}, 332–341.

\bibitem[\protect\citename{Le~Go{\"\i}c {\em et~al.\ }\relax,
  }2016]{LeGoic2016}
Le~Go{\"\i}c, G., Bigerelle, M., Samper, S., Favreli{\`e}re, H., \& Pillet, M.
  2016.
\newblock Multiscale roughness analysis of engineering surfaces: a comparison
  of methods for the investigation of functional correlations.
\newblock {\em Mechanical Systems and Signal Processing}, {\bf 66}, 437–457.

\bibitem[\protect\citename{Ojala {\em et~al.\ }\relax,
  }1996]{ojala1996comparative}
Ojala, Timo, Pietik{\"a}inen, Matti, \& Harwood, David. 1996.
\newblock A comparative study of texture measures with classification based on
  featured distributions.
\newblock {\em Pattern recognition}, {\bf 29}(1), 51--59.

\bibitem[\protect\citename{Ojala {\em et~al.\ }\relax, }2002]{ojala2002outex}
Ojala, Timo, Maenpaa, Topi, Pietikainen, Matti, Viertola, Jaakko, Kyllonen,
  Juha, \& Huovinen, Sami. 2002.
\newblock Outex-new framework for empirical evaluation of texture analysis
  algorithms.
\newblock {\em Pages  701--706 of:} {\em Object recognition supported by user
  interaction for service robots},  vol. 1.
\newblock IEEE.

\bibitem[\protect\citename{Palm \& Lehmann, }2002]{Palm2002}
Palm, C., \& Lehmann, T.~M. 2002.
\newblock Classification of color textures by gabor filtering.
\newblock {\em Machine Graphics and Vision International Journal}, {\bf 11
  (2)}, 195–219.

\bibitem[\protect\citename{Pentland, }1990]{Pentland1990}
Pentland, A.~P. 1990.
\newblock Automatic extraction of deformable part models.
\newblock {\em International Journal of Computer Vision}, {\bf 4}, 107--126.

\bibitem[\protect\citename{Picard, }2000]{VisTex2000}
Picard, R. \&~al. 2000.
\newblock Vision Texture Database.
\newblock {\em http://vismod.media.mit.edu/vismod/imagery/VisionTexture}.

\bibitem[\protect\citename{Pitard, }2016]{pitard2016metrologie}
Pitard, G. 2016.
\newblock {\em M{\'e}trologie et mod{\'e}lisation de l'aspect pour l'inspection
  qualit{\'e} des surfaces}.
\newblock Ph.D. thesis, Universit{\'e} Grenoble Alpes.

\bibitem[\protect\citename{Pitard {\em et~al.\ }\relax, }2017a]{Pitard2017}
Pitard, G., Le~Go{\"\i}c, G., Mansouri, A., Favreli{\`e}re, H., Desage, S.~F.,
  Samper, S., \& Pillet, M. 2017a.
\newblock Discrete Modal Decomposition : a new approach for the reflectance
  modeling and rendering of real surfaces.
\newblock {\em Machine Vision and Applications}, {\bf 28 (5-6)}, 607–621.

\bibitem[\protect\citename{Pitard {\em et~al.\ }\relax,
  }2017b]{pitard2017reflectance}
Pitard, G., Le~Go{\"\i}c, G., Mansouri, A., Favreli{\`e}re, H., Pillet, M.,
  George, S., \& Hardeberg, J.~Y. 2017b.
\newblock Reflectance-based surface saliency.
\newblock {\em Pages  445--449 of:} {\em Image Processing (ICIP), 2017 IEEE
  International Conference on}.
\newblock IEEE.

\bibitem[\protect\citename{Porebski, }2009]{Porebski2009}
Porebski, A. 2009.
\newblock {\em S{\'e}lection d’attributs de texture couleur pour la
  classification d’images : Application à l’identification de d{\'e}fauts
  sur les d{\'e}cors verriers imprim{\'e}s par s{\'e}rigraphie}.
\newblock Ph.D. thesis, Universit{\'e} Lille 1.

\bibitem[\protect\citename{Samper \& Formosa, }2007]{Samper2007}
Samper, S., \& Formosa, F. 2007.
\newblock Form defects tolerancing by natural modes analysis.
\newblock {\em Journal of computing and information science in engineering},
  {\bf 7 (1)}, 44–51.

\bibitem[\protect\citename{Sengur, }2008]{Sengur2008}
Sengur, A. 2008.
\newblock Wavelet transform and adaptive neuro-fuzzy inference system for color
  texture classification.
\newblock {\em Expert Systems with Applications}, {\bf 34 (3)}, 2120–2128.

\bibitem[\protect\citename{Tuceryan \& Jain, }1993]{tuceryanetal1993}
Tuceryan, M., \& Jain, A.~K. 1993.
\newblock Texture analysis.
\newblock {\em Pages  235--276 (Chapter 2) of:} Chen, C.~H., Pau, L.~F., \&
  Wang, P. S.~P. (eds), {\em Handbook of pattern recognition and computer
  vision},  vol. -.
\newblock Singapore: World Scientific Publishing.

\bibitem[\protect\citename{Unser, }1986]{unser1986local}
Unser, Michael. 1986.
\newblock Local linear transforms for texture measurements.
\newblock {\em Signal processing}, {\bf 11}(1), 61--79.

\bibitem[\protect\citename{Vapnik, }1995]{vapnik1995nature}
Vapnik, V.~N. 1995.
\newblock The Nature of Statistical Learning Theory.
\newblock {\em Springer-Verlag New York, Inc.}

\bibitem[\protect\citename{Weber, }1997]{weber1997usc}
Weber, Allan~G. 1997.
\newblock The USC-SIPI image database version 5.
\newblock {\em USC-SIPI Report}, {\bf 315}(1).

\bibitem[\protect\citename{Weston {\em et~al.\ }\relax,
  }1999]{weston1999support}
Weston, J., Watkins, C., {\em et~al.\ }\relax. 1999.
\newblock Support vector machines for multi-class pattern recognition.
\newblock {\em Pages  219--224 of:} {\em Esann},  vol. 99.

\end{thebibliography}






\end{document}